\documentclass[10pt,twocolumn,letterpaper]{article}

\usepackage[pagenumbers]{cvpr} 

\usepackage[dvipsnames]{xcolor}

\definecolor{cvprblue}{rgb}{0.21,0.49,0.74}
\usepackage[pagebackref,breaklinks,colorlinks,citecolor=cvprblue]{hyperref}

\usepackage{graphicx}
\usepackage{amsmath}
\usepackage{amssymb}
\usepackage{booktabs}
\usepackage{algorithmic}
\usepackage{algorithm}
\usepackage{multirow}
\usepackage{xcolor}
\def\paperID{6499} 
\def\confName{CVPR}
\def\confYear{2024}

\title{ LAFS: Landmark-based Facial Self-supervised Learning for Face Recognition }

\author{Zhonglin Sun, Chen Feng\thanks{Corresponding Author.}, Ioannis Patras, Georgios Tzimiropoulos\\
 MMV Group, Queen Mary University of London\\
London, UK\\
{\tt\small \{zhonglin.sun, chen.feng, i.patras, g.tzimiropoulos\}@qmul.ac.uk}
}

\begin{document}
\maketitle
\begin{abstract}

In this work we focus on learning facial representations that can be adapted to train effective face recognition models, particularly in the absence of labels. Firstly, compared with existing labelled face datasets, a vastly larger magnitude of unlabeled faces exists in the real world. We explore the learning strategy of these unlabeled facial images through self-supervised pretraining to transfer generalized face recognition performance. Moreover, motivated by one recent finding, that is, the face saliency area is critical for face recognition, in contrast to utilizing random cropped blocks of images for constructing augmentations in pretraining, we utilize patches localized by extracted facial landmarks. This enables our method - namely \textbf{LA}ndmark-based \textbf{F}acial \textbf{S}elf-supervised learning~(\textbf{LAFS}), to learn key representation that is more critical for face recognition. We also incorporate two landmark-specific augmentations which introduce more diversity of landmark information to further regularize the learning. With learned landmark-based facial representations, we further adapt the representation for face recognition with regularization mitigating variations in landmark positions. Our method achieves significant improvement over the state-of-the-art on multiple face recognition benchmarks, especially on more challenging few-shot scenarios. The code is available at \href{https://github.com/szlbiubiubiu/LAFS_CVPR2024}{https://github.com/szlbiubiubiu/LAFS\_CVPR2024}.


\end{abstract}    
\section{Introduction}
\label{sec:intro}
\begin{figure}[ht]
  \centering
   \includegraphics[width=0.8\linewidth]{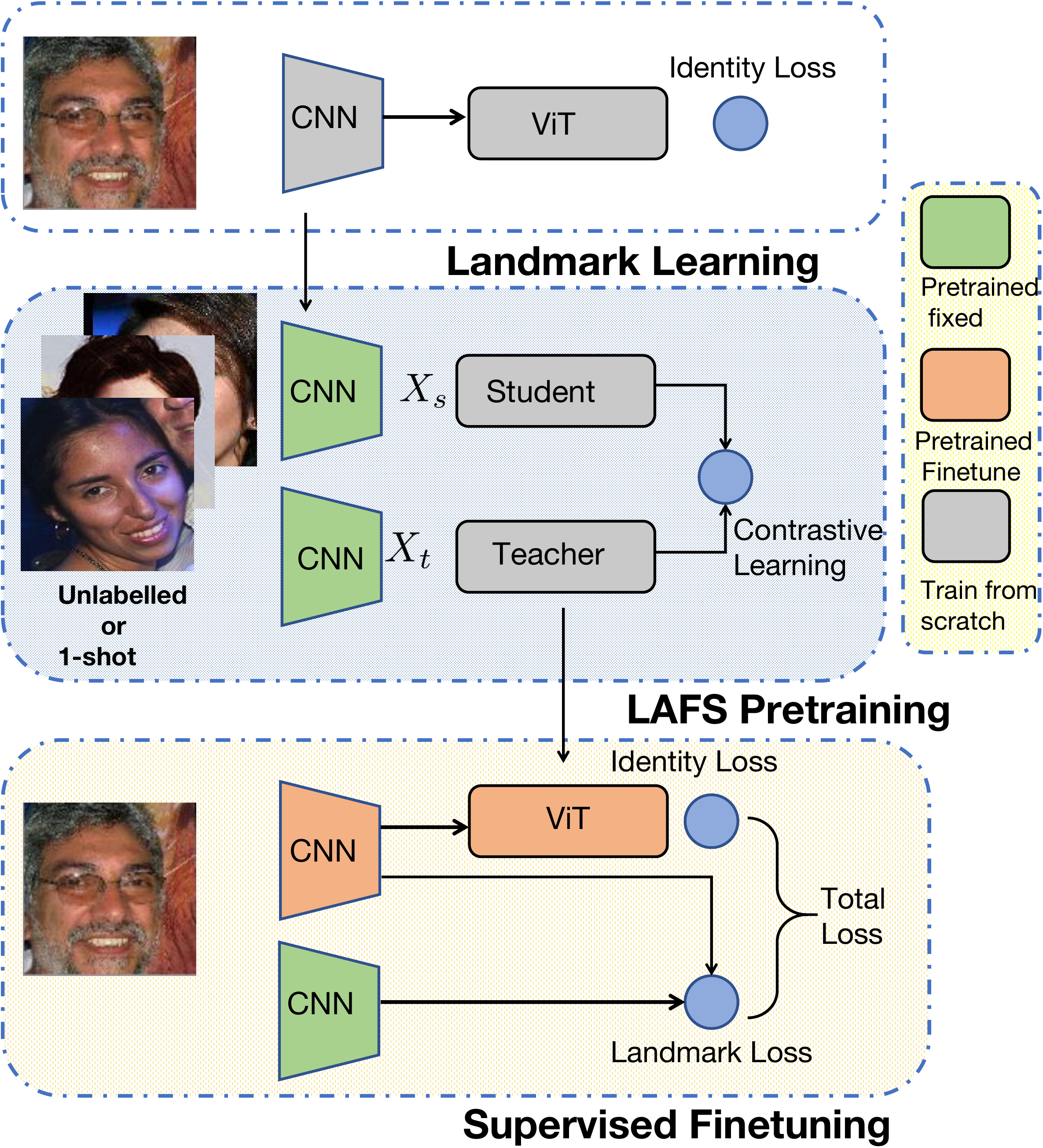}

    \caption{Illustration of our \textit{pretraining} and \textit{finetuning} pipeline for face recognition. First, a landmark CNN is learnt using Part fViT framework~\cite{Sun_2022_BMVC}. We adopt the landmark CNN to provide facial landmarks for constructing our \textbf{LAFS} \textit{pretaining}. In this framework, the `Teacher' processes the entire set of provided landmarks, while the `Student' operates on subsets of these landmarks. Then, we transfer the `Teacher' for \textit{finetuning} with an additional regularization that penalizes landmark predictions from huge variations.}
    
    
   
   \label{fig:pipeline_lafs}
\end{figure}

In recent years, face recognition has witnessed significant advancements, owing to the emerging techniques such as advanced loss functions~\cite{schroff2015facenet,wen2016discriminative,liu2017sphereface,wang2018cosface,arcface,deng2021variational,wen2021sphereface2} and specialized network structures~\cite{chen2013blessing,liu2015targeting,yang2020fan,ding2017trunk,kang2019hierarchical,kang2018pairwise,xie2018comparator,wang2020hierarchical,zhong2021face,Sun_2022_BMVC,dan2023transface}, and the large-scale annotated datasets such as Webface42M~\cite{zhu2021webface260m} and MS1M~\cite{guo2016ms}. However, most of the works have overlooked the impact of initial parameters for supervised training, noted as facial representation, which is commonly proven to be effective. For example, transferring from a good pretrained model derived by either supervised or self-supervised pretraining can yield better results on tasks such as Face Reconstruction~\cite{bulat2022pre}, Face Anti-spoofing~\cite{wang2022patchnet} and ImageNet Classification~\cite{caron2021emerging,xie2022simmim}. A recent empirical facial comparison research~\cite{bulat2022pre} further reveals that self-supervised learning is able to bring much improvement compared to supervised learning, particularly, in data-limited scenarios. However, self-supervised learning with ResNet yields suboptimal results compared to supervised training under large-scale settings~\cite{bulat2022pre}. This observation suggests that previous self-supervised pre-training methods may not effectively scale from limited datasets to scenarios with abundant identity information.

Moreover, another challenge for real-world face recognition is that large-scale datasets are not always available, due to issues such as non-commercial licenses and privacy concerns. Collecting new datasets often ensues in a challenging situation where only a few images are annotated for each novel identity~\cite{yang2023two}. Hence researches under insufficient labelling situation(i.e. few-shot face recognition, especially 1-shot) are worth consideration. Current research on few-shot face recognition focuses primarily on improving accuracy with meta-learning~\cite{holkar2022few,yang2023two,zheng2020novel}, ignoring that a good initial model~(facial representation) is also crucial for few-shot face recognition. A potential solution towards maintaining generalized few-shot accuracy is self-supervised learning~\cite{lu2022self}. 

Motivated by the above concerns, we are interested in three fundamental questions: 
\begin{itemize}
    \item \textit{With a vast number of unlabelled facial images in the real world, how can we take advantage of those data to train a face recognition model?}
    \item \textit{With only a limited number of samples for each identity (few-shot learning), how well can a face recognition model perform and to what extent self-supervised learning can be of improvement over straightforward supervised training?}
    \item \textit{What causes the failure of self-supervised learning when scaling from limited data to large-scale data?}
\end{itemize}

Our paper aims to address the challenge of learning facial representation that can be effectively adapted to face recognition in few-shot and large-scale scenarios. To solve the problems outlined above, we offer 3 contributions:

\begin{itemize}

\item We conduct a series of experiments to answer the question of how well self-supervised learning could behave on few-shot face recognition. We propose a pipeline designed explicitly for few-shot evaluation where we adopt self-supervised methods on an unlabelled(1-shot) pretraining dataset with 1M images which simulate the situation of unlabelled data in real-world applications where each image is treated as a novel class, then we finetune on a few-shot(including 1-shot) datasets and evaluate on the classical test sets(e.g. IJB-C). 


\item Part fViT~\cite{Sun_2022_BMVC} has shown superior localization performance on facial images, indicating that it is feasible to combine a part-based model with self-supervised learning. Inspired by this, We propose a novel landmark-based self-supervised framework for face recognition that pertains entirely to facial parts, where dense landmarks are trained to produce similar representations of sparse landmarks. This facilitates a shift from standard grid learning to landmark-based learning. Fig~\ref{fig:pipeline_lafs} shows the pipeline for our LAFS pretraining and fine-tuning. Furthermore, we investigate the properties of landmarks and propose two landmark-related augmentations to enhance the representation of self-supervised learning.

\item A series of studies are conducted for LAFS, we demonstrate the effectiveness of standard evaluation on large-scale datasets as well as few-shot evaluations, which have never been explored in previous works. 
\end{itemize}
Our main findings are: (a) Without explicit label information, our pipeline, which comprises unlabeled(1-shot) pretraining followed by 1-shot fine-tuning, can deliver accurate face recognition performance. We demonstrate that intra-sample pretraining~(self-supervised learning loss) and inter-sample learning~(1-shot supervised loss, e.g. CosFace~\cite{wang2018cosface}) are both necessary for learning generalized 1-shot face recognition in our pipeline. (b) We find that distinction from the facial research~\cite{bulat2022pre}, self-supervised learning with Vision Transformer can break the limits for self-supervised learning with ResNet, exceeding supervised training by a large margin. (c) Our proposed LAFS, which works entirely on facial landmarks, is capable of transferring to highly accurate face recognition.

\section{Related Work}\label{Sec:related_work}
It is out of scope to review the bulk of papers for face recognition, we introduce the literature for Face Recognition in few-shot and general situations, 
then Self-supervised Learning.

\paragraph{Face Recognition} 
The goal of face recognition is to learn discriminative feature embedding based on the combination of backbone and losses. To this end, some works~\cite{wang2018cosface,deng2019arcface,wen2021sphereface2,liu2016large,kim2022adaface,li2023unitsface} concentrate on modifying the softmax loss function which represents separability to counter the discrimination as softmax loss could not handle the global feature distribution especially when we using mini-batch. CosFace~\cite{wang2018cosface} adds a margin on the $cos(\theta)$ to control the distance between the normalized weight and feature embedding, resulting in a more discriminative feature learning scheme. Following that, ArcFace~\cite{deng2019arcface} proposes that defining the margin on the angle $\theta$ would provide more discriminative results than defining on $cos(\theta)$. In addition, AdaFace~\cite{kim2022adaface} proposes to learn to distinguish low-quality samples and employs adaptive margin to learn more discriminative features. To better extract discriminative representation with loss function, the choice of a suitable backbone plays a pivotal role in computing face images to generate feature embedding. The to-go backbone during the past few years is ResNet~\cite{he2016deep,chen2013blessing,xie2018comparator,yang2020fan,li2023bionet}, recently Vision Transformer are investigated in~\cite{zhong2021face,Sun_2022_BMVC,an2022killing,dan2023transface} for face recognition while training ViT still requires a large amount of data or heavy data augmentation. Part fViT~\cite{Sun_2022_BMVC} is the pioneering study to integrate Vision Transformer with facial parts to learn discriminative facial area which benefits the accuracy. As for the limited data volume, methods for few-shot face recognition~\cite{yang2023two,yu2017discriminative} mostly focus on exploring discriminative features combined with meta-learning training.
\paragraph{Self-Supervised Learning}
Self-supervised learning has achieves great performance as an off-the-shelf representation techniques in various computer vision tasks~\cite{caron2021emerging,chen2020simple,he2020momentum,grill2020bootstrap,caron2020unsupervised,xie2022simmim,he2022masked, feng2022ssr, gao2024self}.
Recently, contrastive learning~\cite{caron2021emerging,chen2020simple,he2020momentum,grill2020bootstrap,caron2020unsupervised, assran2022masked, feng2022adaptive, feng2023maskcon} has made significant strides in exploring image representations based on the instance discrimination task, by considering each sample itself as a unique class. More specifically, two different views of a specific sample are treated as positive to each other while the rest of the dataset is regarded as negative, making the batch size a crucial factor. Large batch sizes are essential for effective training in SimCLR~\cite{chen2020simple}. Instead of contrastive learning, some other works try to avoid instance discrimination tasks. Involving Vision Transformer into consideration, DINO\cite{caron2021emerging} proposes that global augmentation should have similar representations to local augmentations and introduces a centring method to provide stable results.  For face recognition applications, Face-MAE\cite{wang2022facemae} utilizes a masked strategy to reconstruct the face and then minimize the embedding distance between the reconstructed face and the original face. ~\citet{sevastopolskiy2023boost} proposes StyleGAN~\cite{karras2020training} is able to boost FR by reconstructing the input image via taking StyleGan as a self-supervised learning method. 

\begin{figure*}[ht]
  \centering
  \includegraphics[width=0.9\linewidth,height=0.45\linewidth]{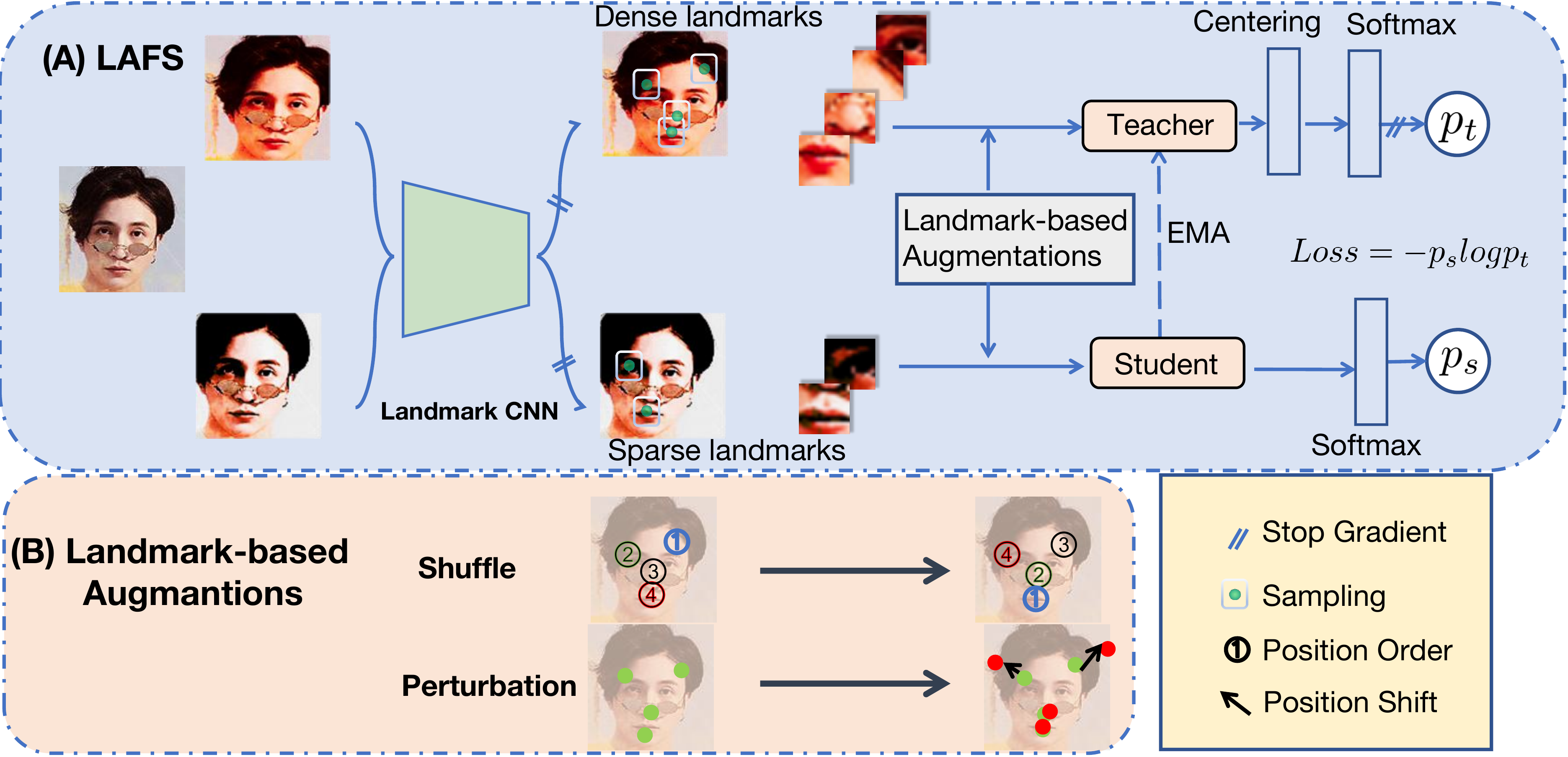}
  \caption{(A)The pipeline of our proposed LAFS framework. Two views of a facial image are first processed by the landmark CNN to provide landmark localization. Then we sample a certain subset of landmarks on the student branch. Following that, landmarks-based augmentations are added before converting into embedding for processing by teacher and student backbones. The representations of the two views are compared by the output of the backbones without label information. Gradients are backpropagated to the student network and the teacher network is updated by the exponential moving average of student parameters. (B)Landmark Augmentations. The upper part is shuffling where the order for sending to fViT is \textcolor{blue}{\textcircled{1}}, \textcolor{green}{\textcircled{2}}, \textcolor{black}{\textcircled{3}}, \textcolor{red}{\textcircled{4}}, after shuffling the order changed. The bottom part explains the coordinates variation given the perturbation, each position of the green point shifts to the red point after the perturbation.}
  \label{fig:LAFS}
\end{figure*}
\section{Methodology}\label{sec:method}


In~\Cref{Sec:DINO_face}, applying DINO~\cite{caron2021emerging} for face recognition pretraining is introduced. Then, a novel self-supervised learning method(LAFS) for face recognition which focuses on the representation between full landmarks and the subset of landmarks is explained in~\Cref {Sec:LAFS}. Furthermore, in \Cref{Sec:Augmentations} we present two types of augmentations. \Cref{sec:few-shot} describes how we simulate using unlabelled data for pretraining. And finally, we explain the finetune strategy for the LAFS in~\Cref{Sec:finetune}. Details for baseline Part fViT~\cite{Sun_2022_BMVC} is described in \textsc{supplementary material}~\ref{sec:landmark_vit}

\subsection{DINO for face recognition}\label{Sec:DINO_face}

We start by adopting the DINO~\cite{caron2021emerging} which provides strong feature embedding through contrastive learning between global views and local views. A facial image X is firstly converted into a pair of views, denoted as $(\mathbf{X_{t}}, \mathbf{X_{s}})$. The objective of the DINO is as follows:
\begin{equation}
    min_{\mathbf{\theta_{s}}} \mathbf{H}(\mathbf{P}(\textrm{Cen}(\mathbf{\vartheta{t}(X_{t})}),\mathbf{\vartheta{s}(X_{s})}))
\end{equation}
Where $\vartheta(*)$ is the backbone, $t$ and $s$ represent the teacher branch and student branch, $\mathbf{P}(*)$ is a projection head with softmax probability, $\textrm{Cen}(*)$ is centring operation which minus the mean computed on the output and $\mathbf{H(a,b)}=-\mathbf{a}log(\mathbf{b})$ is the cross-entropy. DINO endeavours to investigate the interdependency between global and local features in a given image, then the augmentation for the teacher branch is treated as a global representation and the student branch is expected to output a representation of local views. We adopt DINO~\cite{caron2021emerging} as the baseline SSL method for our pipeline. 

\subsection{Full-to-Subset Landmark Consistency Learning}\label{Sec:LAFS}
In contrast to DINO's approach which focuses on the global-local relationship of images, our approach aims to leverage the aggregation and structural characteristics of faces by seeking similar representations between a full set and a subset of facial parts~(landmarks).


\begin{algorithm}

	\renewcommand{\algorithmicrequire}{\textbf{Input:}}
	\renewcommand{\algorithmicensure}{\textbf{Output:}}
	\caption{Pipeline of the proposed Landmark-based Facial self-supervised pretraining}  
	\label{LAFS:algorithm}
	\begin{algorithmic}[1]
            \STATE Initialization: Input image $\mathbf{X}$, teacher network $\mathbf{\vartheta_{t}}$, student network $\mathbf{\vartheta_{s}}$, pretrained landmark CNN $\mathbf{\delta()}$, Landmark Augmentations $\mathbf{\sigma_{l}()}$, DINO augmentations $\mathbf{\sigma_{d}()}$,  moving mean $\mathbf{l}$, $n \leftarrow 0$, maximum iterations $\tau$
            \REPEAT
            \STATE $n \leftarrow n + 1$
            \STATE Sample a batch $\mathbf{X^{n}}$
            \STATE Generate views $\mathbf{X^{n}_{t}}, \mathbf{X^{n}_{s}}\leftarrow \mathbf{\sigma_{d}(X^{n})},\mathbf{\sigma_{d}(X^{n})}$ 
            \STATE $\mathbf{L^{n}_{t}}, \mathbf{L^{n}_{s}}\leftarrow \mathbf{\delta(X^{n}_{t})},\mathbf{Sample(\delta(X^{n}_{s}))}$
            \STATE $\mathbf{\hat{L}_{t}^{n}}, \mathbf{\hat{L}_{s}^{n}}\leftarrow \mathbf{\sigma_{l}(L^{n}_{t})},\mathbf{\sigma_{l}(L^{n}_{s})}$
            \STATE $\mathbf{\hat{L}_{t}^{n}}\leftarrow \mathbf{\hat{L}_{t}^{n}}$.detach()
            \STATE Update $\mathbf{\vartheta_{s}^{n+1}}\leftarrow \mathbf{\vartheta_{s}^{n}} -\nabla_{\mathbf{\vartheta_{s}^{n}}}  $ Eq~\ref{Eq:LAFS}
            \STATE Update $\mathbf{\vartheta_{t}^{n+1}}\leftarrow \mathbf{l}* \mathbf{\vartheta_{t}^{n}}  +\mathbf{(1-l)}\mathbf{\vartheta_{s}^{n}}$
            \UNTIL converges or $\mathbf{n=\tau}$
            \ENSURE output model $\mathbf{\vartheta_{t}^{\tau}}$ and $\mathbf{\vartheta_{s}^{\tau}}$
 
	\end{algorithmic}  
\end{algorithm}

Previous landmark localization methods, such as FAN~\cite{bulat2016human}, are unable to provide landmarks that are specifically relevant for face recognition, thereby negatively affecting the face recognition performance as shown in Part fViT~\cite{Sun_2022_BMVC}. To address this issue, we opt for Part fViT~\cite{Sun_2022_BMVC} for constructing our landmark-based self-supervised learning pipeline. Specifically, as illustrated in (A) part of Figure~\ref{fig:LAFS} and Algorithm~\ref{LAFS:algorithm}, we utilize a landmark CNN $\delta()$, which is fixed and pre-trained on face recognition tasks, to generate a series of facial landmarks before feeding them into the backbone.
\begin{equation}
    \mathbf{L_{t}, L_{s}}=\mathbf{\delta(X_{t})}, \mathbf{\delta(X_{s})}
\end{equation}
This landmark CNN can provide stable and highly aggregated image patches for face recognition. It is crucial to note that if the parameters of this landmark CNN are not fixed, it may converge to the standard ViT grids and negatively impact recognition performance. To establish correspondence between all landmarks and a random subset of landmarks, we randomly sample a portion of landmarks on the student network branch. In our case, we randomly sample 36 out of R=196 global landmarks, which corresponds to the resolution of local views in DINO. 
\begin{equation}
    \mathbf{L_{t}}, \mathbf{L_{s}}=\mathbf{L_{t}},\textrm{Sample}(\mathbf{L_{s}})
\end{equation}
Furthermore, the scale of random crops on the student branch is the same as that of the teacher branch. We append two landmark-based augmentations $\mathbf{\delta_{l}()}$ on the learned landmark to enhance the difficulties of the self-learning. 
\begin{equation}
    \mathbf{\hat{L}_{t}}, \mathbf{\hat{L}_{s}}=\mathbf{\delta_{l}(L_{t})}, \mathbf{\delta_{l}(L_{s})}
\end{equation}
Where $\mathbf{\hat{L}_{t}},\mathbf{\hat{L}_{s}}$ are the full landmark views and subsets of landmark views respectively which can be considered as proprietary versions of the global and local views for face recognition. Our proposed method can effectively explore the representation of global facial landmarks and random subset landmarks. The learning objective for LAFS is the same as that of DINO~\cite{caron2021emerging}:

\begin{equation}\label{Eq:LAFS}
    min_{\mathbf{\vartheta_{s}}} \sum_{\mathbf{\hat{L}_{t}}\in \{ {\mathbf{\hat{L}_{t1}^{g}},\mathbf{\hat{L}_{t2}^{g}}}\} } \sum_{ \mathbf{\hat{L}_{t}}\neq \mathbf{\hat{L}_{s}}} \mathbf{H}(\mathbf{Q_{t}}(\mathbf{\hat{L}_{t}}),\mathbf{Q_{s}}(\mathbf{\hat{L}_{s}})))
\end{equation}

Where $\mathbf{Q_{t}} $ and $\mathbf{Q_{s}}$ are probability produced by the teacher and student backbones, $\mathbf{\hat{L}_{t1}^{g}}$ and $\mathbf{\hat{L}_{t2}^{g}}$ are two global landmark views. $\theta_{s}$ is the student network.

\subsection{Landmark-based data augmentations}\label{Sec:Augmentations}
To further regularize the model, we design two landmark-based data augmentations as shown in (B) part of Figure~\ref{fig:LAFS}, namely Landmark Coordinate Perturbation and Landmark Shuffle. 


\paragraph{Landmark Shuffle}
The landmark CNN, learned from the face recognition task, is capable of finding the corresponding positions of different faces. As a result, the positions of the patches in the part-based model differ from those in the standard ViT, which assumes sequential and non-overlapping positions for patches. The landmarks produced by the landmark CNN do not impose any order constraints, particularly with regard to position, and adjacent landmarks are overlapping as is described in~\cite{Sun_2022_BMVC}. However in vision tasks, maintaining the order of patches preserves the overall structure and global information \cite{naseer2021intriguinga,vaswani2017attention}. In light of this, we hypothesize that strict order is not necessary for the part-based model and have adopted a shuffle operation to perturb the model. Surprisingly, this perturbation significantly improves recognition performance, which contradicts the negative impact of shuffle as observed in \cite{naseer2021intriguinga}. We posit that this efficacy of shuffle on landmarks may be attributed to the differences in position encoding between the Part fViT and standard fViT.


\paragraph{Landmark Coordinate Perturbation}Despite obtaining relatively stable landmarks with the pre-trained and fixed landmark CNN, we suspect that the landmark CNN may not have the same localization performance for all faces, and that the adjacent image regions beyond the landmarks are also worth exploring. Thus we design data augmentation to enable the model to explore the area surrounding the landmarks. Specifically, for given landmarks $r$, we add a coordinate perturbation vector to its coordinate according to the following formula:
\begin{equation}
    \mathbf{r} = \mathbf{r} + \mathbf{\alpha} * \mathbf{u}
\end{equation}

Here, $\mathbf{u}$ is subject to standard normal distribution $\mathbf{u}\sim \mathbf{\mathcal N(0,1)}$, and $\alpha$ controls the magnitude of the disturbance, resulting in a normal distribution with $\mathbf{\mathcal N(0,\alpha^{2})}$.

It is important to note that these two augmentations only show effectiveness in our proposed self-supervised learning method. Therefore, applying these augmentations to the standard backbone, specifically perturbing the DINO+fViT architecture by shuffling, results in a significant decrease in performance. Further details such as appropriate perturbation amplitude and results can be found in Section~\ref{Sec:shuffle} and Section~\ref{Sec:perturbation}.

\subsection{Unlabeled Face Pretraining}\label{sec:few-shot}

We are curious about how well a face recognition model can go with the situation in real-world where images are available without identity information. To address this issue, we simulate by regarding each unlabelled sample as a novel class, resulting in 1-shot face recognition. As a traditional solution, straightforwardly applying classification-based loss such as CosFace~\cite{wang2018cosface} will provide disastrous performance as shown in~\Cref{Aba:finetune_sim_sep}. Inspired by ~\citet{lu2022self} that self-supervised can be a good few-shot learner, we construct a pipeline for few-shot face recognition, where we pretrain on a large-scale unlabelled facial dataset then finetune on few-shot face datasets and finally evaluate the performance on traditional testset. Our experiments make use of the WebFace260M dataset~\cite{zhu2021webface260m}, a clean dataset consisting of 42 million images and 2 million identities, we are able to sample a 1-shot dataset of 1 million classes for our task to simulate the case of real-world application where each image is treated as a new identity. 

\subsection{Finetune}\label{Sec:finetune}
\paragraph{Inter-sample Separability or Intra-sample Similarity}
Self-supervised learning methods primarily focus on exploring the representation of different views of an input image. This raises a pertinent question: When finetuning in a 1-shot setting, which loss function, self-supervised or classification-based, performs better? We address this query through our experiments, wherein we examine applying DINO~\cite{caron2021emerging} loss~(Intra-sample Similarity) and Cosface~\cite{wang2018cosface}~(Inter-sample Separability) for finetuning on the 1-shot dataset in experiment~\ref{Aba:finetune_sim_sep}. DINO loss could not provide superior results compared with Cosface, and it slightly benefits the model without fine-tuning. This indicates that for 1-shot face recognition, simply applying classification loss or self-supervised loss would not benefit the recognition performance, instead, the pipeline unlabeled(1-shot) pretraining with intra-sample similarity~(self-supervised learning) and fine-tuning with inter-sample separability~(Supervised classification) is crucial for addressing the 1-shot problem.

\paragraph{Landmark supervision}\label{sec:land_finetune} We have 3 finetuning options for LAFS finetuning: (a) Fix the landmark CNN and train fViT only; (b) Finetune the whole backbone~(landmark CNN and fViT); (c) As shown in the bottom part of Figure~\ref{fig:pipeline_lafs}, we use an extra pre-trained and fixed landmark CNN to provide landmark localization coordinates as soft-label, and the trainable landmark CNN could be either from the supervised pretraining or self-supervised(e.g. DINO) pretraining. The overall finetuning loss is:
\begin{equation}
    L_{total}=L_{id}+\beta \*||\hat{r},r||_{2}
\end{equation}
Where $\hat{r}$ is the landmarks predicted by the fixed landmark CNN, $r$ is the predicted coordinates, and $\beta$ controls the variation of landmarks. $L_{id}$ is CosFace~\cite{wang2018cosface} to learn discriminative embedding. We ablate the choice of the landmark finetuning options in \textsc{supplementary material}~\ref{Sec:finetune_supp}.


\section{Experiments}\label{sec:experiments}

In this section, we examine the accuracy of models transferring from the proposed LAFS on several well-known datasets as well as the few-shot evaluation, and conduct a series of ablation studies to illustrate the effectiveness of our proposed methods. Text in bold means the best results obtained. We include pretraining, finetuning details as well as the models opted in \textsc{supplementary material}~\ref{sec:supp_exp}.

\subsection{Implementation details}\label{sec:implementation}

To fairly compare with our other methods, for the self-supervised pipeline, we randomly sample 1M images with 1 image per identity from the first 50\% part of the Webface42M dataset for self-supervised pretraining, namely Webface-1shot. As for supervised finetuning, we opt for MS1MV3~\cite{deng2019lightweight} containing 93,431 identities and 10\% of Webface42M~\cite{zhu2021webface260m} dataset, coined as Webface4M, comprising 4M images with 200K identities. We also test how the unlabelled in-the-wild dataset behaves in \textsc{supplementary material}~\ref{sec:implementation_supp}.

\subsection{Ablation Studies}\label{sec:ablation}

A number of studies are conducted to demonstrate the effectiveness of different parts of methods. For self-supervised pretraining, we adopt Part fViT-Tiny with patch number 196, size of 15.28M and 2.48G Flops. We put ablation studies for the number of shots, finetuning landmark supervision, comparison of global landmark view between global view and impact of the in-the-wild dataset in \textsc{supplementary material}~\ref{sec:abl_supp}.

\subsubsection{Finetune with Similarity or Separability}\label{Aba:finetune_sim_sep}

We consider different loss options when transferring from LAFS, specifically the Inter-sample Separability (CosFace~\cite{wang2018cosface} loss) and Intra-sample Similarity (DINO~\cite{caron2021emerging} loss). We report the results of simply adopting the pretrain model for evaluation, then CosFace and DINO in Table~\ref{Tab:dino_cos}. The accuracy for directly training CosFace~\cite{wang2018cosface} on the Webface-1M without pretraining is also included. We can conclude that DINO improve the model without fine-tuning, and CosFace can bring large benefits in terms of accuracy. Then training 1-shot data from scratch leads to invalid results.

\begin{table}[htbp]
\centering
\footnotesize
\resizebox{0.3\textwidth}{!}{%
\begin{tabular}{lccc}
\toprule
Finetune Loss& LFW   & CFP-FP & AgeDB \\ \midrule
w/o Finetune & 75.20 & 70.82  & 55.90 \\ 
DINO         & 76.96 & 73.44  & 56.65 \\ 
CosFace      & 88.58 & 76.04  & 65.84 \\ 
\midrule
CosFace, 1M      & 50.00 & 50.00  & 50.00 \\ \bottomrule
\end{tabular}
}
\caption{Ablation studies for different types of finetune loss. W/o Finetune means directly adopting pretrained model for evaluation. CosFace, 1M means the model is trained from scratch on the 1M 1-shot training set. The backbone adopted is Part fViT-B}
\label{Tab:dino_cos}
\end{table}

\subsubsection{Effect of Landmark Shuffle}\label{Sec:shuffle} 
Herein we challenge the property of the negative impact of shuffle for Vision Transformer which can be found in Table~\ref{Tab:LAFS_Aba} (middle part), as well as the performance of standard fViT with shuffle operations. We can conclude that the shuffling operation indeed disturbs the global information for fViT, but it leads to better recognition performance for Part fViT.

\begin{table}[htbp]
\centering
\resizebox{0.35\textwidth}{!}{%
\begin{tabular}{lcccc}
\toprule
\multirow{2}{*}{Experiment}                                                       & \multirow{2}{*}{Backbone}  & \multirow{2}{*}{Content} & \multirow{2}{*}{1\% data} & \multirow{2}{*}{10\% data} \\
      &                            &                      &                           &                            \\ \midrule
\multirow{4}{*}{Baseline}                                                         
      & \multirow{2}{*}{Part fViT} & w/o ssl              & 6.91                      & 90.87                      \\ \cmidrule(l){3-5} 
      &                            & LAFS w/o Aug         & \textbf{38.05}            & \textbf{91.06}             \\ \cmidrule(l){2-5}
      & \multirow{2}{*}{fViT}      & w/o ssl              & 8.54                      & 86.9                       \\ \cmidrule(l){3-5} 
      &                            & DINO                 & 36.27                     & 87.15                      \\  \midrule 
\multirow{4}{*}{\begin{tabular}[c]{@{}l@{}}Landmark \\ Shuffle\end{tabular}}      & \multirow{2}{*}{\begin{tabular}[c]{@{}c@{}}Part fViT \\ (LAFS)\end{tabular}} & w/o shuffle          & 38.05                     & 91.06                      \\  \cmidrule(l){3-5}  
      &                            & shuffle              & \textbf{40.01}            & \textbf{91.53}             \\ \cmidrule(l){2-5} 
      & \multirow{2}{*}{\begin{tabular}[c]{@{}c@{}}fViT \\ (DINO)\end{tabular}}      & w/o shuffle          & \textbf{36.27}            & \textbf{87.15}             \\ \cmidrule(l){3-5} 
      &                            & shuffle              & 30.96                     & 78.95                      \\ \midrule
\multirow{5}{*}{\begin{tabular}[c]{@{}l@{}}Landmark \\ Perturbation\end{tabular}} & \multirow{3}{*}{\begin{tabular}[c]{@{}c@{}}Part fViT \\ (LAFS)\end{tabular}} & $\alpha$ =0          & 38.05                     & 91.06                      \\ \cmidrule(l){3-5} 
      &                            & $\alpha $=2          & \textbf{39.44}            & \textbf{91.45}             \\ \cmidrule(l){3-5} 
      &                            & $\alpha$ =5          & 39.16                     & 91.37                      \\ \cmidrule(l){2-5} 
      & \multirow{2}{*}{\begin{tabular}[c]{@{}c@{}}fViT \\ (DINO)\end{tabular}}      & $\alpha$ =0          & 36.27                     & 87.15                      \\ \cmidrule(l){3-5} 
      &                            & $\alpha$ =2          & \textbf{36.30}         & \textbf{87.23}             \\ \bottomrule
\end{tabular}}
\caption{Ablation studies for the effectiveness of Landmark augmentations, the top parts are the results for without landmark augmentations. We report IJB-B when TAR@FAR=1e-4}
\label{Tab:LAFS_Aba}
\end{table}

\subsubsection{Effect of Landmark Coordinates Perturbation}\label{Sec:perturbation} We disturb the learned landmark by adding a noise subject to normal distribution. As listed in Table~\ref{Tab:LAFS_Aba} (bottom part), the magnitude with $\alpha=2$ performs the best towards finetuning results. While too large perturbation would even perform worse than without perturbation. 

We also conduct experiments to test the impact of these two augmentations on the standard grid (i.e. fViT). From the results, the following conclusions are observed: (1) Shuffling operation leads to worse results for the standard fViT, which is consistent with previous findings~\cite{naseer2021intriguinga}. (2) Perturbation augmentation could bring small benefits in increasing the performance of the standard fViT. 

\begin{table*}[htbp]
\centering
\resizebox{0.8\textwidth}{!}{%
\begin{tabular}{lcccccccccccc}
\toprule
\multirow{2}{*}{Data Amount} & \multirow{2}{*}{Pretrain Method} & \multirow{2}{*}{Backbone} & \multicolumn{5}{c}{IJB-B}                                                          & \multicolumn{5}{c}{IJB-C}                                                          \\ \cmidrule(l){4-13} 
                             &                                  &                           & 1 shot         & 2 shot         & 4 shot         & 10 shot         & all           & 1 shot         & 2 shot         & 4 shot         & 10 shot         & all           \\ \midrule
\multirow{7}{*}{1\%}         & \multirow{3}{*}{Scratch}         & ResNet                    & 14.13          & 19.69          & 30.58          & 22.19          & 58.66          & 15.39          & 22.15          & 33.56          & 25.01          & 63.16          \\ 
                             &                                  & fViT                      & 1.67           & 3.79           & 6.6            & 16.91          & 27.22          & 1.84           & 4.03           & 7.36           & 19.03          & 29.19          \\ 
                             &                                  & Part fViT                 & 0.64           & 0.89           & 1.46           & 3.47           & 6.22           & 0.58           & 0.97           & 1.59           & 3.73           & 5.99           \\ \cmidrule(l){3-13} 
                             & \multirow{2}{*}{DINO}            & ResNet                    & 13.98          & 22.70          & 36.85          & 52.72          & 72.38          & 15.96          & 26.79          & 41.87          & 57.19          & 77.3           \\ 
                             &                                  & fViT                      & 33.48          & 41.67          & 56.10          & 66.20          & 74.42          & 37.49          & 46.93          & 60.13          & 70.44          & 78.95          \\ 
                             &                                  & Part fViT                 & 18.51          & 23.41          & 35.19          & 49.19          & 69.37          & 21.40          & 27.23          & 39.38          & 53.70          & 75.82          \\ \cmidrule(l){3-13} 
                             & LAFS                             & Part fViT                 & \textbf{33.97} & \textbf{42.80} & \textbf{57.32} & \textbf{68.76} & \textbf{75.67} & \textbf{37.89} & \textbf{47.03} & \textbf{61.09} & \textbf{72.57} & \textbf{79.66} \\ \midrule
\multirow{7}{*}{10\%}        & \multirow{3}{*}{Scratch}         & ResNet                    & 27.23          & 41.96          & 59.27          & 68.25          & 89.33          & 28.84          & 45.25          & 63.26          & 72.40          & 92.28          \\ 
                             &                                  & fViTB                     & 7.53           & 13.32          & 19.85          & 45.45          & 81.37          & 7.63           & 14.21          & 22.11          & 52.80          & 84.62          \\ 
                             &                                  & Part fViT                 & 3.11           & 3.97           & 14.66          & 54.47          & 84.47          & 3.32           & 4.25           & 17.21          & 58.75          & 88.09          \\ \cmidrule(l){3-13} 
                             & \multirow{2}{*}{DINO}            & ResNet                    & 27.63          & 48.97          & 76.83          & 87.58          & 90.99          & 27.91          & 59.58          & 81.29          & 90.26          & 92.28          \\ 
                             &                                  & fViT                      & 46.99          & 66.61          & 81.56          & 88.35          & 91.35          & 51.35          & 70.73          & 85.02          & 91.42          & 93.85          \\ 
                             &                                  & Part fViT                 & 29.82          & 47.48          & 75.92          & 85.39          & 90.37          & 33.87          & 52.45          & 74.60          & 89.07          & 91.44          \\ \cmidrule(l){3-13} 
                             & LAFS                             & Part fViT                 & \textbf{48.56} & \textbf{66.71} & \textbf{81.67} & \textbf{88.70} & \textbf{92.16} & \textbf{51.97} & \textbf{71.10} & \textbf{85.09} & \textbf{91.56} & \textbf{94.32} \\ \bottomrule
\end{tabular}}
\caption{The comparison of the proposed methods on few-shot evaluation}
\label{Tab:fewshot}
\end{table*}

\subsection{Comparison with the State-of-the-Art}
We experiment with ResNet-100, fViT-B and Part fViT-B models to evaluate the performance transferring from the self-supervised pipeline on well-known datasets to compare with the recent state-of-the-art methods as well as the results fine-tuned with few-shot samples.


The training~(or fine-tuning) datasets we used for quantitative results are Webface4M~\cite{zhu2021webface260m}, few-shot subsets of Webface4M and MS1MV3~\cite{guo2016ms}. We achieve SOTA or near SOTA results on the Webface4M and MS1MV3 dataset, and our proposed LAFS pretraining outperforms other methods for few-shot evaluation. Note that those results are finetuned with (c) mentioned in Section~\ref{sec:land_finetune}. 

\subsubsection{Few-shot Evaluation}
We report the performance few-shot evaluation on a few subsets of Webface4M~\cite{zhu2021webface260m} in Table~\ref{Tab:fewshot}, specifically 1\% labels with randomly 1, 2, 4, 10 shots and all images per label, and 10\% labels with 1, 2, 4, 10 shots and 10\% labels with their available images. The evaluation is performed on three backbones: ResNet-100, fViT-B, and Part fViT-B, considering both the supervised training and the self-supervised pretraining settings. We also test the impact of the in-the-wild dataset Flickr which is collected by us in \textsc{supplementary material}~\ref{sup:flickr}. Results for Part fViT with fixed landmark CNN under DINO training are also provided. Based on the evaluation results, we can draw the following conclusions:

\begin{itemize}
    \item Without self-supervised pretraining, we consistently observe that ResNet outperforms both fViT and Part fViT in few-shot evaluation scenarios. This gap can be attributed to the overfitting property of Vision Transformer when trained on limited amounts of data~\cite{zhong2021face,steiner2021train}.
    \item Part fViT can exceed fViT when a sufficient volume of data is available under the supervised training setting, specifically when the data amount is equal to or larger than 10\% label with 10 shots.
    \item Self-supervised learning consistently yields superior results compared to training without self-supervised training, except for the ResNet under 1\% label with 1-shot setting. Notably, the benefits of self-supervised learning become more prominent when the volume of available data decreases, which aligns with the observations from the facial research~\cite{bulat2022pre}.

    \item Consistent with the findings of DINO~\cite{caron2021emerging}, we observe that the improvement achieved through self-supervised learning for ResNet is not as significant as that observed for fViT in most cases. 
    \item Our proposed LAFS pretraining could always obtain effective results compared with models pre-trained by DINO. Particularly, when the number of shots reaches the maximum available data, we observe significant improvements over DINO which aligns with supervised finding, implying that more diversity in each identity will help with finding effective landmarks. However, DINO with Part fViT would provide worse training results compared to fViT with DINO, which illustrates the effectiveness of our LAFS pipeline.
\end{itemize}

\begin{table}[ht]
\centering
\resizebox{0.9\linewidth}{!}{%
\begin{tabular}{lcccccc}
\toprule
Method                                    & SSL  & LFW   & CFP-FP   & AgeDB & IJB-B          & IJB-C          \\ \midrule
ArcFace w/o Aug~\cite{zhu2021webface260m} & -    & 99.85 & 99.04 & 97.82 & -              & 96.77          \\
CosFace w/o Aug~\cite{zhu2021webface260m} & -    & 99.80 & 99.25 & 97.45 & -              & 96.86          \\
QAFace~\cite{saadabadi2023quality}        & -    & 99.85 & 99.21 & 97.91 & 95.67          & 97.20          \\
AdaFace~\cite{kim2022adaface}             & -    & 99.80 & 99.17 & 97.90 & 96.03          & 97.39          \\
ArcFace PFC-0.3~\cite{an2022killing}      & -    & 99.85 & 99.23 & 98.01 & 95.64          & 97.22          \\ \midrule
ResNet, w/ Aug                           & -    & 99.83 & 99.16 & 97.89 & 95.94          & 97.34          \\
ResNet, w/ Aug                           & DINO & 99.82 & 99.14 & 97.9  & 95.83          & 97.29          \\
fViT~\cite{Sun_2022_BMVC}                                      & -    & 99.83 & 99.04 & 97.48 & 95.72          & 97.15          \\
fViT                                      & DINO & 99.83 & 99.10 & 97.77 & 95.91          & 97.33          \\
Part fViT~\cite{Sun_2022_BMVC}                                 & -    & 99.83 & 99.13 & 97.73 & 96.05          & 97.4           \\
Part fViT                                 & LAFS & 99.83 & 99.18 & 97.90 & \textbf{96.28} & \textbf{97.55} \\ \midrule
fViT, w/o Aug                             & DINO & 99.85 & 99.05 & 97.70 & 95.64          & 97.12          \\
Part fViT, w/o Aug                        & LAFS & 99.81 & 99.14 & 97.83 & \textbf{96.07} & \textbf{97.44} \\ \bottomrule
\end{tabular}%
}
\caption{Comparison with the state-of-the-art on Webface4m. LAFS outperform other methods on several testsets.}
\label{Res:Webface_SOTA}
\end{table}

\subsubsection{Webface4M} We present the results of the proposed self-supervised pretraining on the 1-shot image set, then finetuning on Webface4M in Table~\ref{Res:Webface_SOTA}. We also test the results of ResNet-100 with augmentations as well as ResNet-100 pretrained from DINO. We also self-implement Part fViT on Webface4M with the same hyper-parameters of~\cite{Sun_2022_BMVC} for comparison as it is not reported.


\begin{table*}[htbp]
\centering
\resizebox{0.8\textwidth}{!}{%
\begin{tabular}{lccccccccc}
\toprule
Method                                 & SSL  & Train Data & LFW            & CFP-FP         & AgeDB          & IJB-B          & IJB-C          & MegaFace/id    & MegaFace/ver   \\ \midrule
CosFace\cite{wang2018cosface}          & -    & MS1MV2     & 99.81          & 98.12          & 98.11          & 94.80          & 96.37          & 97.91          & 97.91          \\
ArcFace\cite{deng2019arcface}          & -    & MS1MV2     & 99.83          & 92.27          & 92.28          & 94.25          & 96.03          & 98.35          & 98.48          \\
Sub-center ArcFace\cite{deng2020sub}   & -    & MS1MV2     & 99.80          & 98.80          & 98.31          & 94.94          & 96.28          & 98.16          & 98.36          \\
FAN-Face\cite{yang2020fan}             & -    & MS1MV2     & 99.85          & 98.63          & 98.38          & 94.97          & 96.38          & 98.70          & 98.95          \\
VirFace\cite{li2021virtual}            & -    & MS1MV2     & 99.56          & 97.15          & -              & 88.90          & 90.54          & -              & -              \\
MagFace\cite{meng2021magface}          & -    & MS1MV2     & 99.83          & 98.46          & 96.15          & 94.51          & 95.97          & -              & -              \\
Face Transformer\cite{zhong2021face}   & -    & MS1MV2     & 99.83          & 96.19          & 97.82          & -              & 95.96          & -              & -              \\ \midrule
ArcFace-challenge\cite{deng2021masked} & -    & MS1MV3     & 99.85          & 99.06          & 98.48          & -              & 96.81          & -              & -              \\
VPL\cite{deng2021variational}          & -    & MS1MV3     & 99.83          & 99.11          & \textbf{98.60} & 95.56          & 96.76          & 98.80          & \textbf{98.97} \\
AdaFace\cite{kim2022adaface}           & -    & MS1MV3     & 99.83          & 99.03          & 98.17          & 95.84          & 97.09          & -              & -              \\ \midrule
ResNet, w/ Aug                        & -    & MS1MV3     & 99.85          & 99.14          & 98.16          & 96.03          & 97.22          & 98.78          & 98.67          \\
ResNet, w/ Aug                        & DINO & MS1MV3     & 99.85          & 99.12          & 98.14          & 95.86          & 97.01          & 98.50          & 98.53          \\
fViT~\cite{Sun_2022_BMVC}              & -    & MS1MV3     & 99.85          & 99.01          & 98.13          & 95.97          & 97.21          & 98.69          & 98.91          \\
fViT, ours                             & DINO & MS1MV3     & 99.85          & 99.08          & 98.10          & \textbf{96.13} & 97.27          & 98.70          & 98.65          \\
Part fViT~\cite{Sun_2022_BMVC}         & -    & MS1MV3     & 99.83          & \textbf{99.21} & 98.29          & 96.11          & 97.29          & \textbf{98.96} & 98.78          \\
Part fViT, ours                        & LAFS & MS1MV3     & \textbf{99.86} & 99.15          & 98.33          & \textbf{96.24} & \textbf{97.48} & 98.66          & \textbf{98.95} \\ \midrule
fViT, w/o Aug, ours                    & DINO & MS1MV3     & 99.81          & 99.02          & 98.10          & 95.76          & 97.14          & 98.15          & 98.31          \\
Part fViT, w/o Aug, ours               & LAFS & MS1MV3     & 99.83          & 99.11          & 98.33          & \textbf{95.98} & \textbf{97.20} & 98.73          & 98.62          \\ \bottomrule
\end{tabular}%
}
\caption{Comparison with the state-of-the-art on MS1MV3 datasets. Our proposed LAFS achieve state-of-the-art results on most datasets.}
\label{SOTA}
\end{table*}

It can be observed on large pose evaluation provided by the CFP-FP protocol, Part fViT boosts the fViT by 0.09 while producing similar results with ResNet-100 under the same training augmentations. Self-supervised methods bring small improvement, such that LAFS outperforms Part fViT by 0.05 and DINO with fViT improves supervised fViT by 0.06. However, for age variation evaluation provided by the AgeDB-30 dataset, the self-supervised learning proves advantageous for all considered backbones, our LAFS improves Part fViT by 0.17, fViT with DINO improves fViT by 0.29, ResNet with DINO improves ResNet by 0.01. And LAFS still surpasses baseline fViT with DINO which is 0.13. On IJB-B and IJB-C datasets, baseline fViT could produce competitive results compared with AdaFace~\cite{kim2022adaface}, and the proposed LAFS is able to achieve much higher accuracy by 0.16 on IJB-C and 0.25 on IJB-B than AdaFace. In terms of improvement of SSL, SSL could generate promising improvement (i.e. DINO improves fViT by 0.18 and LAFS boosts Part fViT by 0.15 on IJB-C).


In addition, we also provide results trained without data augmentation for a fair comparison with the previous setting, our LAFS outperforms the SOTA method AdaFace~\cite{kim2022adaface} by 0.04 on IJB-B and 0.05 on IJB-C. fViT with DINO behaves slightly worse than ArcFace PFC-0.3~\cite{an2022killing} and QAFace~\cite{saadabadi2023quality} which is 95.64 on IJB-B and 97.12 on IJB-C, but still a large increase compared to supervised CosFace~\cite{wang2018cosface}. Note that AdaFace is trained with cropping, scaling, and jitting augmentations. Another observation is that self-supervised learning could only provide marginal improvement or even worse for ResNet, which is consistent with the facial research~\cite{bulat2022pre}.

\subsubsection{MS1MV3}~\label{sec:MS1MV3}
As the original version of MS1MV*, MS-Celeb~\cite{guo2016ms}, is abandoned by the creator, AdaFace~\cite{kim2022adaface} advocates for researchers to transition towards using the Webface4M~\cite{zhu2021webface260m}, hence for the sake of reference, we report the results of the models trained on MS1MV3, and test on various benchmarks. The results are shown in Table~\ref{SOTA}. 


As observed, fViT with DINO pretraining is able to boost the accuracy, especially on IJB-B and IJB-C datasets by a large margin over supervised fViT, e.g. 0.16 on IJB-B and 0.06 on IJB-C. Remarkably, the performance achieved by LAFS for Part fViT considerably outweighs that of DINO, achieving the best performance on MS1MV3 which improves Part fViT-B without self-supervised learning by 0.13 on IJB-B and 0.19 on IJB-C. Also LAFS outperforms the state-of-the-art method AdaFace~\cite{kim2022adaface} by 0.4 and 0.39 respectively. 


In addition, regarding results without strong data augmentation, our LAFS outperforms the SOTA method AdaFace~\cite{kim2022adaface} by 0.14 on IJB-B and 0.11 on IJB-C. And for ResNet-100 the same phenomenon is observed with SSL~\cite{bulat2022pre} that with 100\% of the dataset, self-supervised learning is not capable of bringing improvement when the data volume is large, it even results in a reduction in accuracy, for instance, dropping from 97.22 with supervised training to 97.01 with SSL pretraining.

\section{Conclusions}\label{sec:conclusion}
We propose landmark-based self-supervised learning (LAFS) to learn generalized representations that can be adapted to highly accurate face recognition. We offer several key contributions: (A) A self-supervised pipeline utilizing un-labelled data for training generalized representation, which can be adopted to evaluate the effectiveness of self-supervised representation. (B) LAFS, a novel self-supervised learning framework called landmark-based self-supervised learning for face recognition, minimises the representation of all landmarks and fewer landmarks to provide promising improvement when transferring to face recognition. (C) Two augmentations, namely Landmark Shuffle and Landmark Coordinates Perturbation, are effective and robust for learning more generalized representations for LAFS. Overall, our approach achieves state-of-the-art or near state-of-the-art performance on several face recognition benchmarks including few-shot evaluation, demonstrating the effectiveness of our proposed methods. However, the landmarks in LAFS are fixed and don't benefit from self-supervised training. In the future, we will explore more techniques to refine the landmarks during self-supervised training stages. In addition, we will investigate to what extent the LAFS can behave on other facial tasks beyond face recognition.
\paragraph{Acknowledgments:} This research utilised Queen Mary's Apocrita HPC facility, supported by QMUL Research-IT and ITS Research.

%


{
    \small
    \bibliography{main}
    \bibliographystyle{ieeenat_fullname}
}

\clearpage
\setcounter{page}{1}
\maketitlesupplementary

\section{Introduction}
This is the supplementary material for the paper \textbf{LAFS: Landmark-based Facial Self-supervised Learning for Face Recognition}. We first introduce the baseline Part fViT in Section~\ref{sec:landmark_vit}. Then we describe the dataset collection details, model details, and hyper-parameters details for DINO, LAFS and fine-tuning in Section~\ref{sec:implementation_supp}, Section~\ref{model_details} and Section~\ref{sec:hyper_details} respectively. Finally, additional ablation studies including the number of shots, finetuning options for landmark supervision and comparison of full landmark view and global view are given in Section~\ref{sec:abl_supp}. 


\section{Additions to section~\ref{sec:method}: Method}\label{sec:supp_method}

\subsection{Part fViT}\label{sec:landmark_vit}

Face saliency area has shown its capability of improving the recognition accuracy~\cite{chen2013blessing}. In this work, we adopt Part fViT~\cite{Sun_2022_BMVC} as our default backbone for recognition and investigate the property of the learned landmark CNN. Part fViT consists of two sequential components:(a) A landmark CNN which is responsible for predicting patch (landmark) centres and extracting corresponding patches. (b) A patch-based backbone, namely ViT, to predict the final identity embedding. 

An image $X$ is processed by a light-weight CNN to compute $R$ landmark center $r$, 
\begin{equation}
\mathbf{r} = \textrm{CNN}(\mathbf{X}),\; r_i=[x_i,y_i]^T,\; i=1,\dots,R
\end{equation}
Where $r$ is the coordinates representing the landmark centre, normalized by the min-max scaler. Then, a patch whose centre is given by the landmark coordinate $r_i$ with a fixed size of 8 in $R=196$ is sampled using the differentiable grid sampling method of STN~\cite{jaderberg2015spatial}. Following this, each patch is projected by the linear layer $\mathbf{E}$, then positioned by the positional encoding, appended with the class tokens before feeding to Transformer.


The common pipeline for ViT is non-overlapped patch split, patch to embedding projection with positional embedding, and Transformer block consists of self-attention and feed-forward networks. For more comprehensive information, please refer to~\cite{dosovitskiy2021an}.  

The derivative of the Part fViT, the landmark CNN, is of providing stable landmarks with good correspondence~\cite{Sun_2022_BMVC}, giving us a hint that it can be useful in developing a new perspective for self-supervised learning for face recognition.

\begin{table*}[t]
\centering
\resizebox{0.8\textwidth}{!}{%
\begin{tabular}{@{}lcccccccccccc@{}}
\toprule
\multirow{2}{*}{Data amount} & \multirow{2}{*}{Pretrain Method} & \multirow{2}{*}{Backbone} & \multicolumn{5}{c}{IJB-B}                  & \multicolumn{5}{c}{IJB-C}                  \\ \cmidrule(l){4-13} 
                             &                                  &                           & 1 shot & 2 shot & 4 shot & 10 shot & full  & 1 shot & 2 shot & 4 shot & 10 shot & full  \\ \midrule
\multirow{5}{*}{1\%}         & Scratch                          & Resnet                    & 14.13  & 19.69  & 30.58  & 22.19   & 58.66 & 15.39  & 22.15  & 33.56  & 25.01   & 63.16 \\ \cmidrule(l){3-13} 
                             & DINO                             & fViT                      & 33.48  & 41.67  & 56.10  & 66.20   & 74.42 & 37.49  & 46.93  & 60.13  & 70.44   & 78.95 \\ \cmidrule(l){3-13} 
                             & LAFS                             & Part fViT                 & \textbf{33.97}  & \textbf{42.80}  & \textbf{57.32}  & \textbf{68.76}   & \textbf{75.67} & \textbf{37.89}  & \textbf{47.03}  & \textbf{61.09}  & \textbf{72.57}   & \textbf{79.66} \\ \cmidrule(l){3-13} 
                             & DINO(flickr-ours)                & fViT                      & 17.23  & 20.65  & 28.34  & 47.67   & 69.38 & 21.43  & 23.59  & 32.98  & 47.32   & 73.11 \\ \cmidrule(l){3-13} 
                             & LAFS(flickr-ours)                & Part fViT                 & 17.55  & 21.72  & 30.79  & 47.76   & 70.11 & 22.10  & 25.95  & 34.89  & 50.63   & 73.85 \\ \midrule
\multirow{5}{*}{10\%}        & Scratch                          & Resnet                    & 27.23  & 41.96  & 59.27  & 68.25   & 89.33 & 28.84  & 45.25  & 63.26  & 72.40   & 92.28 \\ \cmidrule(l){3-13} 
                             & DINO                             & fViT                      & 46.99  & 66.61  & 81.56  & 88.35   & 91.35 & 51.35  & 70.73  & 85.02  & 91.42   & 93.85 \\ \cmidrule(l){3-13} 
                             & LAFS                             & Part fViT                 & \textbf{48.56}  & \textbf{66.71}  & \textbf{81.67}  & \textbf{88.70}   & \textbf{92.16} & \textbf{51.97}  & \textbf{71.10}  & \textbf{85.09}  & \textbf{91.56}   & \textbf{94.32} \\ \cmidrule(l){3-13} 
                             & DINO(Flickr-ours)                & fViT                      & 28.87  & 44.39  & 71.51  & 76.11   & 89.15 & 31.55  & 57.01  & 80.17  & 79.86   & 85.76 \\ \cmidrule(l){3-13} 
                             & LAFS(Flickr-ours)                & Part fViT                 & 29.07  & 45.67  & 71.93  & 76.56   & 90.90 & 31.98  & 57.65  & 80.31  & 80.71   & 86.74 \\ \bottomrule
\end{tabular}}
\caption{Ablation study for different pretraining dataset. The default pretraining dataset is Webface-1shot, and Flickr-our denotes the Flickr data we collected.}
\label{tab:flickr}
\end{table*}

\section{Additions to section~\ref{sec:experiments}: Experiments}\label{sec:supp_exp}
\subsection{ Implementation details}

\subsubsection{Dataset Details}\label{sec:implementation_supp}

Images are aligned by~\cite{deng2019arcface} and resized to 112$\times$ 112. Our models are evaluated on LFW~\cite{huang2008labeled}, CFP-FP~\cite{sengupta2016frontal}, AgeDB-30~\cite{moschoglou2017agedb}, IJB-B\cite{whitelam2017iarpa}, IJB-C\cite{maze2018iarpa} and MegaFace\cite{kemelmacher2016megaface}. We report 1:1 verification accuracy on  LFW, CFP-FP and AgeDB-30 datasets. Performance of Tar@Far=1e-4 is reported for IJB-B and IJB-C datasets. For Megaface, we report rank-1 identification accuracy (\%) on 1M distractors and TAR@FAR=1e-6 verification accuracy, noted as Megaface/id and Megaface/ver respectively. We adopt a series of data augmentation while they are not typically used for Resnet training settings~\cite{deng2019arcface,wang2018cosface}. We also provide results without data augmentation.

\paragraph{Flickr Dataset} The Webface4M dataset are curated dataset containing high-quality and clean faces, we follow facial research~\cite{bulat2022pre} to collect Flickr dataset in which images are collected from in-the-wild. Differes from facial research~\cite{bulat2022pre}, we downloaded images with following tags: \textit{40s, 50s, 60s, 70s, 80s, 90s, baby, boss, celebrity, face, human}. All images are normalized and aligned by Retina-face~\cite{deng2020retinaface}. In total, we collect 1.2M images which provide a similar data volume to our pretraining Webface 1-shot dataset. 



\subsubsection{Model details}\label{model_details}
We use fViT as proposed in Part fViT~\cite{Sun_2022_BMVC} as the default backbone with comparable parameters and Flops to ResNet-100~\cite{he2016deep}. The landmark CNN is pretrained from Part fViT~\cite{Sun_2022_BMVC} which is MobilenetV3~\cite{howard2019searching}.



\subsubsection{Details for self-supervised pretraining and finetuning}\label{sec:hyper_details}

\paragraph{DINO pretraining}\label{Sec:DINO_setting}
We follow the bulk of the default setting in DINO~\cite{caron2021emerging}, specifically includes 2 global views and 8 local views, the augmentations contain Gaussian Blur, ColorJitter, Random Grayscale, and Solarization. The learning rate is 5e-4 and the optimizer is AdamW~\cite{loshchilov2017decoupled}. The exception is that the number of epochs is 40 while the warmup epoch is 10, the resolution of the global view is set to be 112 and for the local view is 48, and the dimensionality of the DINO head is 100K.

\paragraph{Landmark-based pretraining}
Our LAFS pretraining starts by taking a fixed landmark CNN pre-trained from supervised learning. Since the augmentation for supervised training and self-supervised pretraining is different, the landmark CNN is not capable of producing accurate localization for self-supervised~(DINO) augmented images. To address this, we disentangle the augmentations for the landmark CNN, while maintaining the flip and random resize \& crop operations, in order to generate landmarks for the augmented images. Next, we change the local crop scale to be the same as the global views, i.e. from $[0.08-0.4]$ to $[0.4,1.0]$. After converting the image into patches which are set to be R=196, we randomly sample 36 out of the 196 landmarks before feeding them to the student branch.

\paragraph{Finetuning}

We follow Part fViT~\cite{Sun_2022_BMVC} to use the same regularization and data augmentations. Then We adopt layer-wise learning rate decay~\cite{singh2015layer} of 0.58 inspired by SimMIM~\cite{xie2022simmim}. The optimizer opted for is AdamW~\cite{loshchilov2017decoupled}. We conduct finetuning for 34-80 epochs based on the amount of available data, where we use 34 epochs for 100\% of the data and 80 epochs for 1\% of the data. The weight decay for all networks is 1e-1, the learning rate is 1e-4 with 5 warm-up epochs and cosine learning rate decay.

\subsection{Ablation Study}\label{sec:abl_supp}

\subsection{Impact of In-the-wild dataset}\label{sup:flickr}
We ablate the choice of pretraining data in this part, where we use Flickr and Webface-1shot for pretraining. Results are presented in Table~\ref{tab:flickr}. As it is shown, regardless of the amount of data, adopting Flickr as a pretraining dataset consistently has a negative impact. It even performs worse than Resnet pretrained by DINO on the 1shot setting using Webface-1shot. However, even with this negative impact, using Flickr as a pretraining dataset still yields better results compared to training the model from scratch.

\begin{table*}[ht]
\centering
\footnotesize

\resizebox{0.8\textwidth}{!}{%
\begin{tabular}{@{}l|l|cccc|ccc@{}}
\toprule
\multirow{2}{*}{Experiment}                   & \multirow{2}{*}{Content}      & \multicolumn{4}{c}{1\% data}                            & \multicolumn{3}{c}{100\% data}                   \\ \cmidrule(l){3-6} \cmidrule(l){7-9} 
                                              &                           & LFW   & CFP-FP & AgeDB & IJB-B                          & CFP-FP & AgeDB & IJB-B                           \\ \midrule
\multirow{2}{*}{Number of shot}               & 4-shot pretraining        & 87.96 & 71.7   & 66.08 & 40.75                          & 95.70  & 95.28 & 91.64                           \\ \cmidrule(l){2-6} \cmidrule(l){7-9} 
                                              & 1-shot pretraining         & 88.5  & 72     & 66.9  & \textbf{41.3} & 95.82  & 95.57 & \textbf{91.90} \\ \midrule
\multirow{6}{*}{Finetune options}             & (1) Fixed landmark            & 87.34 & 70.93  & 64.67 & 38.31                          & 95.14  & 95.06 & 87.82                           \\ \cmidrule(l){2-9} 
                                              & (2) Trainable landmark        & 88.53 & 72.19  & 66.8  & 41.25                          & 95.94  & 95.51 & 91.85                           \\ \cmidrule(l){2-9} 
                                              & (3) $\beta=$100                     & 87.37 & 70.88  & 64.53 & 38.24                          & 95.10  & 95.10 & 87.87                           \\ \cmidrule(l){2-9} 
                                              & (3) $\beta=$1                       & 88.01 & 71.39  & 65.37 & 39.47                          & 95.34  & 95.27 & 90.14                           \\ \cmidrule(l){2-9} 
                                              & (3) $\beta=$0.1                     & 88.5  & 72     & 66.9  & \textbf{41.3} & 95.82  & 95.57 & \textbf{91.90} \\ \cmidrule(l){2-9} 
                                              
                                              & (4) landmark to standard grid & 90.06 & 72.57  & 63.56 & 21.53                          & 93.44  & 91.68 & 86.60                           \\ \midrule
\multirow{3}{*}{Global view vs Landmark View} & Global views              & 86.7  & 70.3   & 64.9  & 39.5                           & 95.32  & 95.22 & 91.54                           \\ \cmidrule(l){2-9} 
                                              & Mixed views               & 88.1  & 71.2   & 65.6  & 40.7                           & 95.68  & 95.29 & 91.71                           \\ \cmidrule(l){2-9} 
                                              & Landamrk View             & 88.5  & 72     & 66.9  & \textbf{41.3} & 95.82  & 95.57 & \textbf{91.90} \\ \bottomrule
\end{tabular}}
\caption{Ablation studies for the number of shots, finetuning options and impact of different views. }
\label{Tab:LAFS_ABA}
\end{table*}
\subsubsection{Effect of number of shots}\label{Sec:fewshot}

We conduct experiments to show how the number of shots in pretraining affects the fine-tuning results. To this end, we construct a training set consisting of 1 million images with 250k identities, which we refer to as the 4-shot setting. We compare pretraining on this 4-shot dataset to pretraining on the 1-shot setting, and find that pretraining on the 1-shot dataset yields better results, as is shown at the top of Table~\ref{Tab:LAFS_ABA}.

\subsubsection{Effect of Landmark Finetuning choices}\label{Sec:finetune_supp}
In this study, we aim to investigate the tolerance of fViT to varying the patch coordinates. To this end, we design four fine-tuning methods: (1) fixing the landmark CNN, (2) training the entire landmark CNN and fViT, (3) using an additional pre-trained and fixed landmark CNN to provide pseudo labels, and then training the entire network,  We also exam (4) self-supervised training under the landmark pattern, then finetuning with a standard grid. Moreover, under the setting of (3), we control the strength of the pseudo-label to supervise the new model, that is, $\beta=0.1, 1, 100$, 
results are available in the middle part of Table~\ref{Tab:LAFS_ABA}.

Our experimental results show that the strong supervision of $\beta=100$ gives similar results to (2), while (4) results in worse training results than training fViT from scratch. We can conclude that when the gap of the input grid is very large, self-supervised pre-training and fine-tuning may lead to invalid pre-training. On the other hand, weak supervision of $\beta=0.1$ achieves the best recognition accuracy. This indicates that adhering strictly to the supervised pattern of landmark coordinates will not obtain the optimal result. These findings suggest that Vision Transformers~(Part fViT and fViT) are sensitive to coordinates~(grids), and exploring the surrounding area of the landmark (i.e., using coordinate perturbation) during the pre-training stage may provide a more general landmark CNN for face recognition in the fine-tuning stage.

\subsubsection{Full Landmark View vs Global View}
Our idea for this work is to minimise the presentation of all landmarks (full landmark views) with subsets of landmarks. However, we also explore the possibility of requiring the global image to have a similar representation of a subset of landmarks. To investigate this, we design an experiment where we compared the performance of our method with varying numbers of landmark views on the teacher branch as is shown in Table~\ref{Tab:LAFS_ABA} (bottom part). Mixing views means the teacher processes global view(standard grid) and landmark view at the same time. The results indicate that with more landmark views on the teacher branch, finetuning performance behaves better. It can be drawn that minimising the representation with the global view with subset landmarks is much more challenging, and training with landmark view only can produce better self-supervised pretraining representation.
\begin{table}[htbp]
\centering
\begin{tabular}{|l|l|l|}
\hline
\multirow{2}{*}{Pretraining} & \multirow{2}{*}{IJB-B} & \multirow{2}{*}{IJB-C} \\
                             &                        &                        \\ \hline
WebFace4m 1-shot             & 48.56                  & 51.97                  \\ \hline
MS1M Random                  & 48.47                  & 51.76                  \\ \hline
\end{tabular}
\caption{Difference between 1-shot and unlabeled pretraining}
\label{supp:unlabel_1shot}
\end{table}
\subsubsection{1-shot Simulate unlabeled problem}
We are aware that 1-shot pretraining may not accurately replicate the real-world conditions. To tackle this issue we randomly select 1M facial images from the MS1MV3 dataset to compare the difference between 1-shot pretraining and unlabeled pretraining. Finetune is carried on 1-shot with 10\% of the available data as outlined in Table~\ref{supp:unlabel_1shot}. One can conclude that unlabeled pretraining brings slightly worse but comparable performance than that of 1-shot pretraining.

\end{document}